\documentclass[lettersize,journal]{IEEEtran}
\usepackage{amsmath,amsfonts}
\usepackage{algorithmic}
\usepackage{algorithm}
\usepackage{array}
\usepackage[caption=false,font=normalsize,labelfont=sf,textfont=sf]{subfig}
\usepackage{textcomp}
\usepackage{stfloats}
\usepackage{url}
\usepackage{verbatim}
\usepackage{graphicx}
\usepackage{cite}
\usepackage[utf8]{inputenc} 
\usepackage[T1]{fontenc}    
\usepackage{hyperref}       
\usepackage{url}            
\usepackage{booktabs}       
\usepackage{amsfonts}       
\usepackage{nicefrac}       
\usepackage{microtype}      
\usepackage{xcolor}         
\usepackage{multicol}
\usepackage{xspace}
\usepackage{multirow}
\usepackage{listings}
\usepackage{amsmath,amsfonts}
\usepackage{amssymb}
\usepackage{algorithmic}
\usepackage{algorithm}
\usepackage{booktabs}
\usepackage{array}
\usepackage{textcomp}
\usepackage{stfloats}
\usepackage{url}
\usepackage{verbatim}
\usepackage{graphicx}
\usepackage{colortbl}
\usepackage{color}
\usepackage{subcaption}
\usepackage{makecell}
\usepackage[flushleft]{threeparttable}
\usepackage{orcidlink}
\usepackage{enumitem}
\usepackage{xcolor}
\usepackage{soul}        
\soulregister\cite7
\newcommand{\ours}{DexH2R\xspace}
\begin{document}

\title{DexH2R: Task-oriented Dexterous Manipulation from Human to Robots}

\author{Shuqi Zhao$^{*}$, Xinghao Zhu$^{*}$, Yuxin Chen, Chenran Li, Yichen Xie, Xiang Zhang, Mingyu Ding$\dag$, Masayoshi Tomizuka
\thanks{*Equal Contribution. $\dag$Corresponding author. All authors are with the Department of Mechanical Engineering, the University of California, Berkeley. }}

\markboth{Journal of \LaTeX\ Class Files,~Vol.~14, No.~8, August~2021}%
{Shell \MakeLowercase{\textit{et al.}}: A Sample Article Using IEEEtran.cls for IEEE Journals}

\IEEEpubid{0000--0000/00\$00.00~\copyright~2021 IEEE}

\maketitle

\begin{abstract}
Dexterous manipulation remains a challenging problem due to the high-dimensional action space of robot hands. To solve this challenge, recent advances focus on transferring human hand motions to dexterous robot motions. However, the most straightforward method, retargeting, fails to ensure successful task execution due to the ignorance of hand-object interaction and environmental feedback, while other methods combining retargeting with human feedback significantly increase human labor and degrade motion quality. To address this limitation, we propose a novel framework that enhances retargeted motions with a residual policy based on reinforcement learning, allowing robots to autonomously correct action errors without human intervention. Specifically, our approach leverages human demonstrations to provide natural and task-relevant finger motions while incorporating object and environmental information to improve manipulation performance. Extensive experiments in multiple robot hand embodiments demonstrate that our method outperforms baseline approaches by approximately 40\% in task success rates on both seen and unseen objects. Furthermore, our approach eliminates the need for real-time human correction, significantly reducing labor costs and facilitating scalable data collection for dexterous manipulation. 
\end{abstract}


\section{Introduction}
\IEEEPARstart{H}{umans} manipulate a wide range of objects with their hands in both industrial processes and daily life. This extraordinary ability is also desirable for dexterous robot hands to handle various objects in diverse manipulation tasks. However, further development in this field is hindered by the difficulty in the generation of successful trajectories for dexterous manipulation given the high dimensional action space of dexterous robot hands.
%
%
To solve this challenge, human motions serve as an easily accessible guidance of dexterous manipulation controller in the wild. Therefore, recent advancement has focused on transferring human hand motions to dexterous robot hand motions~\cite{qin2022dexmv, chen2022dextransfer}, which leads to ample development in retargeting.
%
As a promising way, retargeting~\cite{qin2023anyteleop, handa2020dexpilot} aims at maximizing the kinematics similarity between human hand and robot hand, which is defined by the relationship between their keypoints. However, retargeting alone often fails to yield effective manipulation outcomes in practice due to the ignorance of manipulation results like object information or self-collision of dexterous robot hand.
%
%
For example, when grasping a mug, retargeting only maps the keypoints from human hand to robot hand, but doesn't consider whether the mug is actually lifted and following the desired trajectory at all.
%

One common approach to address such problem is combining retargeting and human feedback together~\cite{wang2024dexcap}.
\IEEEpubidadjcol
By doing so, human can iteratively correct the manipulation errors after performing retargeting at each step based on the feedback from environments, which mitigates the negative impact of retargeting errors in a close-loop manner~\cite{qin2023anyteleop, cheng2024open}.
\begin{figure}[t]
  \centering
   \includegraphics[width=0.9\columnwidth]{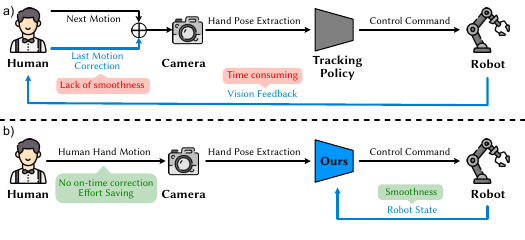}
   \includegraphics[width=0.9\columnwidth]{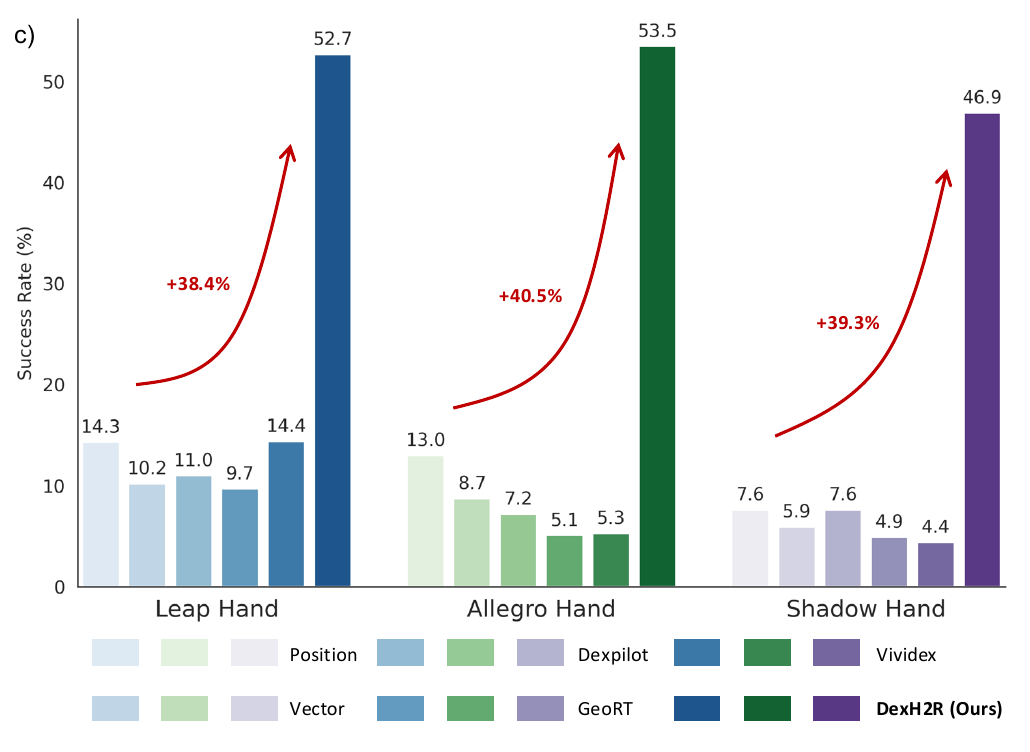}
   \caption{Unlike traditional methods with human feedback\cite{gao2022efficient,si2024tilde,qin2022one} (Fig. a), \ours operates without real-time human intervention ((Fig. b)), significantly reducing human effort, ensuring smooth operation and outperforming baselines by approximately 40\% on different robot hands (Fig. c)).}
   \label{fig:teaser}
\end{figure}
However, incorporating human feedback introduces two additional challenges:
First, the reliance on human feedback makes manipulation process highly time-consuming and labor-intensive.
Second, human tend to make sudden corrective movements in response to system errors, leading to unnatural and non-smooth robot motions.
These two disadvantages significantly reduce the transferring efficiency and degrade the quality of robot motions compared to actual human behaviors.

To this end, we propose a novel method that automatically compensates for action errors by leveraging reinforcement learning, thus gets rid of the need for human feedback.
Specifically, we develop a residual module to correct the robot hand motions generated by the retargeting module. Taking both environment and object information as inputs, this task-oriented residual model remains beneficial for the improvement of concrete manipulation performance rather than simply mapping the human hand motions.
As illustrated in Figure~\ref{fig:teaser}, our method bridges the gap between human hand and dexterous robot hand, enables robots to successfully finish the trajectories without any human correction, thus eliminating the high cost of human labor and time consumption. Moreover, our method is compatible with different embodiments and can be directly applied on different robot hands, free from heavy hyper-parameter tuning or human hand manipulation dataset adjustment. 
Such benefits enable our method to obtain robot manipulation data with high quality in different kinds of dexterous robot hands, which can potentially serve as a scalable way to collect a large amount of dexterous robot data.
Overall, contributions of our work are summarized as follows:
\begin{itemize}
    \item We~\cite{cheng2025open}  propose a novel human hand translator for dexterous manipulation, which enables dexterous hands to imitate human hand motions and finish the manipulation tasks at the same time. This algorithm can be generic and compatible with different robot hands.
    \item We develop a residual action policy based on reinforcement learning to boost the primitive actions generated through retargeting by leveraging both environment and object information.
    \item We conduct extensive experiments to evaluate our method on three different embodiments using both seen and unseen objects. It notably outperforms baselines that purely consider human hand motions by approximately 40\% and demonstrates feasibility in real-world experiments.
\end{itemize}

\section{Related Work}
\label{sec:rw}
\subsection{Human-robot Motion Alignment}
\noindent Due\cite{sitilde}\cite{liudextrack}\cite{she2022learning}\cite{wang2023mimicplay}\cite{sivakumar2022robotic}\cite{citation-0}\cite{luo2024omnigrasp}\cite{romero2017embodied}\cite{makoviychuk2021isaac}\cite{NEURIPS2018_7634ea65}\cite{Rajeswaran-RSS-18} to the structural similarity between human and dexterous hands, human hand motions are considered as a good guidance of dexterous robot hand motions. Many previous works aim to obtain dexterous hand motions from human hand motions. The most straightforward method is retargeting~\cite{antotsiou2018task,arunachalam2023dexterous,chen2023genaug}, which aligns keypoints positions~\cite{qin2023anyteleop} or vectors~\cite{handa2020dexpilot, qin2023anyteleop} between human hands and robot hands to kinematically maximize their similarity. To further improve the transferring performance, human feedback is often introduced to provide on-time correction in a closed-loop manner based on the feedback from environments. Specifically, visual-based teleoperation receives visual feedback from third-view~\cite{gao2022efficient,liu2022robot}, in-hand~\cite{si2024tilde,li2020mobile,li2022dexterous}, ego-centric camera~\cite{qin2022one,li2019vision,aronson2022gaze}, VR~\cite{zhang2018deep,ponomareva2021grasplook,cheng2024open,arunachalam2023holo,lipton2017baxter}, human eyes~\cite{yin2025dexteritygen} and force-based teleoperation requires operators to wear special equipment to receive real-time force feedback from robot hands~\cite{liu2017glove}. With sufficient environmental feedback, operators can adjust the hand gesture accordingly, thus ensuring the success task completion. However, the performance and efficiency of methods with human feedback are both limited due to the cost and intensity of human labor. 

Compared to aforementioned approaches, our method takes task completion into consideration and requires no human correction when transferring human motions to robot motions, thus improving both performance and efficiency.

\subsection{Dexterous Hand Manipulation}
\noindent Dexterous manipulation remains a big challenge due to its high-dimensional action space~\cite{mania2018simple} and complicated mechanical structures among different types of dexterous hands~\cite{shaw2023leap, shadow,allegro,si2024deltahands} for diverse tasks such as pick-and-place, in-hand rotation, and bi-manual manipulation~\cite{rajeswaran2017learning,chen2022system,huang2021generalization,gupta2021reset,huang2021generalization,chen2022towards,zhao2025dexctrl}. As a result, directly applying a specially designed controller or calculating dexterous actions optimization can be limited and high-cost, thus leading to the exploration of learning methods which obtains better potential in generalization.
Reinforcement Learning (RL) is widely introduced as a concise task-oriented approach~\cite{rajeswaran2017learning, zhu2019dexterous,li2025maniptrans, khandate2023sampling,liu2025dextrack,she2022learning,nagabandi2020deep}. Focusing on grasping objects and following targeted object trajectories~\cite{bae2023pmp, wang2023physhoi,braun2024physically}, PGDM~\cite{dasari2023learning} trains individual policy for each trajectory and emphasizes pre-grasp as a crucial feature in the object grasping phase. By leveraging real robot data, GraspGF learns a score-based grasping primitive action and trains a residual policy through RL as a task-oriented finetune~\cite{wu2023graspgf}. Besides RL, Imitation Learning (IL) is also explored in dexterous hand~\cite{argall2009survey, englert2018learning,wang2023mimicplay,wan2023unidexgrasp++,fang2025anydexgrasp}. By leveraging human videos\cite{sivakumar2022robotic,bharadhwaj2024towards,qin2022dexmv,shaw2024learning} or extracted human hand trajectories\cite{wang2024dexcap, chen2022dextransfer, gu2023rt, xu2023xskill}, IL shows relatively convincing performances with less real robot data. Furthermore, \cite{chen2024vividex} and \cite{luo2024grasping} provide a reinforcement learning policy that enables objects to follow targeted trajectories after retrieving primitive actions through human demonstration. 

Different from their methods, ours focuses on translating human motions to dexterous hand behaviors to realize successful manipulation by leveraging robot-object interaction information through simulation, thus replicating human hand motions as closely as possible without any requirements on real robot data. Table~\ref{table:rw} demonstrates the differences between \ours and other work.
\begin{table}[t]
\caption{Overview of differences between our method and related work.}
\small
\setlength{\tabcolsep}{10pt}
\begin{threeparttable}
\centering
\resizebox{\columnwidth}{!}{
\begin{tabular}{c|cccccc}
\toprule
  & \makecell{Human hand\\guidance} & \makecell{Generalize in \\one policy} & \makecell{Imitate \\human motions}
  & \makecell{No human \\ correction}
    & \makecell{Doesn't need \\robot data}\\ 
\midrule
Retargeting~\cite{yin2025geometric,qin2023anyteleop,handa2020dexpilot} & $\checkmark$ & $\checkmark$ & $\checkmark$ & $\times$ & $\checkmark$\\
 Dexcap~\cite{wang2024dexcap} & $\times$ & $\checkmark$ & $\checkmark$ & $\times$ & $\checkmark$\\
 Dexmv~\cite{qin2022dexmv} & $\checkmark$ & $\checkmark$ & $\times$ & $\checkmark$ & $\checkmark$\\
 GraspGF~\cite{wu2023graspgf} & $\checkmark$ & $\checkmark$ & $\times$ & $\checkmark$ & $\times$ \\
 PGDM~\cite{dasari2022learning} & $\times$ & $\times$ & $\times$ & $\checkmark$  & $\checkmark$ \\
 ViVidex~\cite{chen2025vividex} & $\checkmark$ & $\times$ & $\times$ & $\checkmark$ & $\checkmark$ \\
 \textbf{Ours} & \textbf{$\checkmark$} & \textbf{$\checkmark$} & \textbf{$\checkmark$} & \textbf{$\checkmark$} & \textbf{$\checkmark$} \\

\bottomrule     
\end{tabular}
}
\end{threeparttable}
\label{table:rw}
\end{table}
\section{Method}
\label{sec:method}
\begin{figure*}[t]
  \centering
   \includegraphics[width=0.9\textwidth]{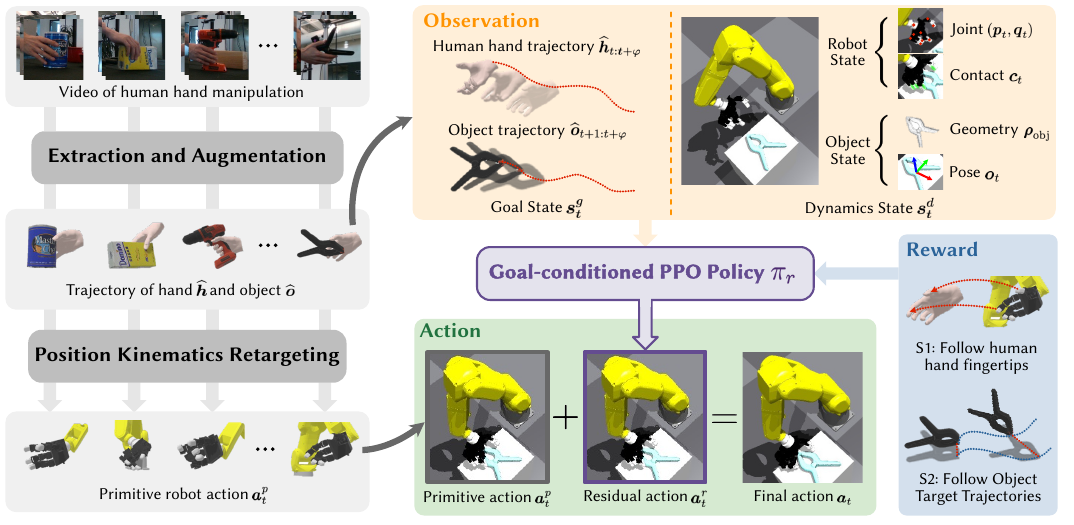}
   \caption{Overview of \ours framework. After extracting and augmenting human hand and object information from demonstrations, we obtain a large amount of trajectories distributed over all workspace. We perform position kinematics retargeting to acquire primitive actions $\boldsymbol{a}_t^p$. Afterwards, taking in both human hand and object trajectories, we learn a residual action $\boldsymbol{a}_t^r$ to equip our method with task completion information. The primitive and residual actions are combined together as the final actions.}
   \label{fig:overview}
\end{figure*}
\noindent Illustrated in Figure~\ref{fig:overview}, our method that transfers human hand actions to robot hand actions consists of two key stages: acquiring primitive actions through retargeting and obtaining residual actions via reinforcement learning. In the first stage, we first enhance the diversity of human demonstrations via data augmentation. Then, the primitive actions are derived through retargeting optimization based on human hand motions, which provides a dexterous grasping solution at the kinematic level. In the second stage, we employ reinforcement learning to develop a residual policy to polish these primitive actions and produce the final actions. 

Overall, our two-stage method delivers a series of task-accomplished actions transferred from human actions by taking in human demonstration trajectories as well as the current robot state and object state.
Leveraging the residual policy to compensate lack of environment and object information when generating primitive actions through retargeting, our method can effectively transfer human hand motions to dexterous hand motions with successful task completion, narrowing the gap between human and robot hands.
\subsection{Preparing Human Data: Extraction and Augmentation}
\label{section:method_dataset}
\noindent To perform hand poses extraction from grasping images, we first identify the hand poses and keypoints using the MANO model~\cite{romero2022embodied}, which derives hand keypoints from human hand images. To improve the generalization ability of policy with a limited set of human hand demonstrations, we perform data augmentation on the poses of both hands and objects~\cite{chen2022dextransfer}. Specifically, $\boldsymbol{\tau} = \{(\boldsymbol{h}_1, \boldsymbol{o}_1), (\boldsymbol{h}_2, \boldsymbol{o}_2), \dots, (\boldsymbol{h}_N, \boldsymbol{o}_N)\}$  represents a trajectory composed of human hand poses $\boldsymbol{h}_t$ and object poses $\boldsymbol{o}_t$. Here, human hand poses $\boldsymbol{h}_t$ are represented as several human hand keypoints positions extracted by MANO. Then, the augmented trajectory $\boldsymbol{\tau}^{'}$ is derived as follows:
\begin{gather}
    \boldsymbol{h}_t^{'} = \boldsymbol{\mathcal{T}}^{'} \boldsymbol{h}_t, \ \boldsymbol{o}_t^{'} = \boldsymbol{\mathcal{T}}^{'} \boldsymbol{o}_t, \\
    \boldsymbol{\tau}^{'} = \left\{\left(\boldsymbol{h}_1^{'}, \boldsymbol{o}_1^{'}\right),\ldots,\left(\boldsymbol{h}_t^{'}, \boldsymbol{o}_t^{'}\right),\ldots,\left(\boldsymbol{h}_N^{'}, \boldsymbol{o}_N^{'}\right)\right\},
\end{gather}
where $\boldsymbol{\mathcal{T}}^{'}$ is the transformation matrix. It translates the trajectories in 3D space and rotates them along the gravity direction. The augmentation parameters are chosen based on two criteria: (i) the object must lie within the robot’s workspace, and (ii) the object should be distributed as broadly as possible across that workspace. The procedure is as follows.
1) We first sample object poses uniformly within the workspace and record the transformation matrix $\boldsymbol{\mathcal{T}}^{'}$ from each sampled pose to the original dataset pose.
2) We then apply $\boldsymbol{\mathcal{T}}^{'}$ to the robot end-effector pose in Cartesian space, which preserves the relative pose between the robot and the object and thus maintains physical feasibility.
3) If the transformed end-effector pose admits a valid inverse-kinematics solution, we compute the corresponding joint angles and check temporal smoothness against recent history.

Such criteria ensures the consistent and physically plausible hand–object relative motion, and further promotes a more uniform distribution of the augmented dataset over the workspace. In this case, we generate additional feasible demonstration trajectories with diverse object poses and reduce the requirement on the amount of human data, which is crucial for robust policy training.

\subsection{Primitive Actions: Kinematic Retargeting}
\label{sec:primitive}
\noindent To facilitate policy training and minimize redundant exploration, we initialize the policy with retargeting solutions. The retargeting process maps human hand motions to robot’s hand joints based on human demonstrations. However, due to different sizes and shapes of human and robot hands, direct keypoint mapping~\cite{qin2023anyteleop, shaw2023leap} can result in infeasible robotic motions. To address this, we exert a scaling factor $\alpha$ to human manipulation trajectory to approximate the size of robot hand. This scaling factor is computed based on the ratio between fingertip span of the robot and that of the human, ensuring approximate spatial alignment between corresponding keypoints. Subsequently, we formulate a non-linear optimization problem to minimize the topological discrepancies between the keypoints of human and robot hands as follows:
\begin{gather}   
    \min_{\boldsymbol{q}_t} \sum_{k=0}^{N}  \left\|\alpha\left(\boldsymbol{h}_t^k-\boldsymbol{o}_{p,t}\right)-\left(f_k(\boldsymbol{q}_t)-\boldsymbol{o}_{p,t}\right)\right\|^2,     \\
    \text{s.t. }  \|\boldsymbol{q}_t-\boldsymbol{q}_{t-1}\| \leq \boldsymbol{d}.
\end{gather}
where $\boldsymbol{q}_t$ represents the robot joint angles at time step $t$. $\boldsymbol{o}_{p,t}$ is the position of the object.  $\boldsymbol{h}_t^k,k=1,2,\dots,N$  denotes the position of  the $k$-th human hand keypoint. The function $f_k(\boldsymbol{q}_t)$ computes the position of the $k$-th robot hand keypoint through forward kinematics. $\boldsymbol{d}$ is a threshold for joint angles difference between adjacent frames. 
$N$ and $\alpha$ are fixed per embodiment. $N$ denotes the number of hand keypoints used. With two keypoints per finger, we have $N = 2 \times n_{\text{fingers}}$; for example, Allegro and Leaphand each have four fingers, giving $N=8$. $\alpha$ is the robot--human size scale, i.e., $\alpha = \text{(robot hand length scale)} / \text{(human hand length scale)}$. We follow standard values in practice: $\alpha = 1.6$ for Allegro/Leaphand and $\alpha = 1.2$ for Shadow. The spatial step $d$ is determined by the joint maximum velocity $v_{\max}$ and the retargeting loop frequency $f_z$ as $d = v_{\max}/f_z$. In practice, we set $d$ slightly larger than $v_{\max}/f_z$ to accommodate minor timing jitter and avoid oscillations.

After solving the optimization, we define the primitive action as $\boldsymbol{a}_t^p = \boldsymbol{q}_t-\boldsymbol{q}_{t-1}$ to represent the delta values of joint angles, which are used along with the residual actions.

\subsection{Residual Actions: Goal-conditioned Reinforcement Learning}
\noindent While the primitive action $\boldsymbol{a}_t^p$ provides a solution for manipulation, its performance can be highly unstable in real-world deployment since retargeting focuses solely on mapping human hand movements to the robot hand without considering the specific task goals. This open-loop system lacks human correction that exists in previous works~\cite{he2024learning, cheng2024open, qin2023anyteleop}, so kinematic retargeting errors can accumulate along the trajectory. We observe that the retargeting performance is significantly affected by the gap between human and robot hand, keypoint detection inaccuracies and control errors during deployment, which often causes task failures.

To this end, we introduce a residual policy to refine the final actions by evaluating whether the objects are following the desired trajectories. By incorporating task-specific feedback, the policy compensates for errors in both keypoint extraction and retargeting, leading to improved stability and success rates during manipulation. Specifically, the residual policy is developed based on reinforcement learning whose state and reward are explained as below.

\textbf{State} The state $\boldsymbol{s}_t$ at each time step consists of the dynamic state $\boldsymbol{s}_t^d$ and goal state $\boldsymbol{s}_t^g$:
\begin{gather}
    \boldsymbol{s}_t \triangleq (\boldsymbol{s}_t^d, \boldsymbol{s}_t^g),\\
    \boldsymbol{s}_t^d \triangleq (\boldsymbol{\rho}_{\text{obj}}, \boldsymbol{o}^{\text{all}}_t,\boldsymbol{q}_t,\,\,\boldsymbol{p}_t-\boldsymbol{o}_{p,t},\,\boldsymbol{c}_t),\\
    \boldsymbol{s}_t^g \triangleq (\widehat{\boldsymbol{h}}_{t:t+\varphi}, \widehat{\boldsymbol{o}}^{\text{all}}_{t+1:t+\varphi}-\boldsymbol{o}^{\text{all}}_t),
\end{gather}
\begin{figure}[t]
  \centering
   \includegraphics[width=0.7\columnwidth]{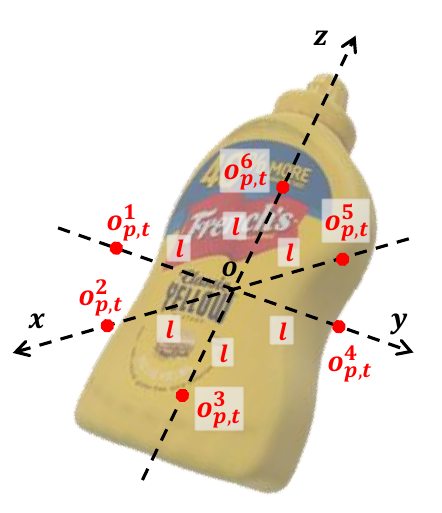}
   \caption{Illustration of keypoints $\{\boldsymbol{o}_{p,t}^{i}\}^6_{i=1}$ that are used to define object poses.}
   \label{fig:object}
\end{figure}

The dynamic state $\boldsymbol{s}_t^d$ represents the states decided only by the environment (including object and robot states). $\boldsymbol{q}_t$ is current joint angles. $\boldsymbol{o}_t^{\text{all}} = \{\boldsymbol{o}_{p,t}^{i}\}^6_{i=1}$ stands for positions of six keypoints decided by the object coordinate frame, representing the objects poses. As shown in Figure~\ref{fig:object}, for each principal axis of the object coordinate frame ($x$, $y$, and $z$), we select two points that lies at a fixed distance $l$ from the object center along both the positive and negative direction of that axis. $\boldsymbol{p}_t-\boldsymbol{o}_{p,t}$ emphasizes the relative positions between robot joints and objects. We use a binary variable $\boldsymbol{c}_t$ to represent the contact state between each robot finger and object due to the inaccuracy and significant sim-to-real gap of contact forces measurement in simulation. $\boldsymbol{\rho}_{\text{obj}}$ denotes the object information including its scale and shape, enabling \ours to be generalizable among different objects. The goal state $\boldsymbol{s}_t^g$ reflects the desired future states of both the human hands $\boldsymbol{\hat{h}}_{t:t+\varphi}$ and object $\boldsymbol{\hat{o}}^{\text{all}}_{t+1:t+\varphi}$ between frame $t$ (or $t+1$) and $t+\varphi$ extracted from human demonstrations, which guides current policy with future prospect to reduce ambiguity.

\textbf{Reward.} 
We propose a two-stage reward function to guide the manipulation process with different goals in each stage: (1) Following the human demonstration when the robot hand is away from the object, and (2) Tracking the object trajectory when the robot hand is close to the object. Concretely, the reward function is written as:
\begin{equation}
    r_t = \left\{ 
\begin{array}{ll}
r_t^{\text{hand}} & \text{if } t \leq t_0 \\
r_t^{\text{obj}} & \text{if } t > t_0 
\end{array} \right.\\    
\end{equation}

When the robot hand is far from the object, $r_t^{\text{hand}}$ encourages the robot hand to move closer to the object. This reward is designed to compensate for the inaccuracies of primitive actions caused by inevitable errors in optimization computation and robot control.
Once the robot hand is close enough, $r_t^{\text{obj}}$ makes the object follow the targeted trajectory, which is the primary objective of the manipulation task. We empirically set a pre-defined switching time $t_0$ to transit between these two stages, which is 15 steps before the object is lifted in demonstrations. The detailed reward function for each stage is shown as follow:
\begin{gather}
r_t^{\text{hand}} = \beta_{\text{hand}}*\exp\left(-\gamma_{\text{hand}}\textstyle\sum_{m=0}^{4} \, ||\widehat{\boldsymbol{h}}_t^m - \boldsymbol{p}_t^m||^2\right),\\
\begin{gathered}
r_t^{\text{obj}} = \beta_{\text{obj}}^{\text{close}}*\exp\left(-\gamma_{\text{obj}}^{\text{close}}\textstyle\sum_{m=0}^{4} \, ||\boldsymbol{p}_t^m - \boldsymbol{o}_{p,t}||^2\right)\\
+ \beta_{\text{obj}}^{\text{follow}}*\exp\left(-\gamma_{\text{obj}}^{\text{follow}}||\widehat{\boldsymbol{o}}^{\text{all}}_t - \boldsymbol{o}^{\text{all}}_t||^2\right)
\end{gathered}
\end{gather}
where $m$ represents the index for each fingertip and $\beta,\gamma$ are hyperparameters. This staged reward function brings a smooth and accurate manipulation process from approaching the object to completing the task.

\section{Experiments}
\label{sec:experiment}
\subsection{Experiment Setup}
\noindent \textbf{Demonstration Dataset.}
For our human demonstration dataset, we utilize DexYCB~\cite
{chao2021dexycb}, which focuses on grasping tasks. It contains 20 objects in total, with approximately 25 grasping trajectories for each object using diverse grasp poses. In our study, we leave out 4 objects for unseen test objects while the remaining objects are adopted in training. Due to the inherent differences between human hands and the hardware of dexterous robot hands, we filter the dataset based on two key criteria:
1) The trajectory must be achievable using only rigid body contact, without relying on deformable materials to provide additional friction or forces from multiple directions (\textit{e.g.} tasks like lifting an upside-down bowl). 2) The object should be placed stably without the risk of falling or rolling away during manipulation.
We further expand the dataset using the approach detailed in Section~\ref{section:method_dataset} to generate demonstrations that cover the entire workspace. Additionally, we interpolate the robot joint trajectories to match its limited velocity constraints for robots.

\textbf{Baselines.}
We compare our method against three widely-used retargeting techniques that aim to transfer human hand motions to robot motions:
\begin{itemize}[leftmargin=10pt]
    \item \textbf{Position Retargeting}~\cite{qin2023anyteleop}: This method minimizes the distances between the keypoints of the human hand and the robot hand. It is exactly the same approach that generates the primitive actions in our method (Sec.~\ref{sec:primitive}).
    \item \textbf{Vector Retargeting}~\cite{qin2023anyteleop}: This technique aligns the vectors from the wrist to the fingertips between human and robot embodiments. It also calculates the robot arm joint angles using IK based on the relative poses between the human wrist and the robot's initial frame.
    \item \textbf{Dexpilot Retargeting}~\cite{handa2020dexpilot}: Similar to vector retargeting, this method aligns both the vectors from the wrist to the fingertips and the vectors between fingertips for each embodiment. The arm joint angles are also computed using IK.
    \item \textbf{GeoRT}~\cite{yin2025geometric}:  GeoRT tackles retargeting by training a model with a designed loss that maps human finger keypoints to robot hand keypoints, achieving high speed and accuracy.
    \item \textbf{ViVidex-SP}~\cite{chen2025vividex}:  ViViDex-SP denotes the state policy of ViViDex. Like our method, it uses RL with trajectory-guided rewards to learn state-based policies from human motions. However, it (1) trains a separate policy per trajectory, while our single model handles all trajectories; (2) does not take human-hand motions as input, so it misses fine-grained hand cues; and (3) relies on precomputed retargeting only up to pre-grasp, whereas we condition directly on human-hand keypoints and use retargeted motions merely as primitives. Consequently, ViViDex-SP has to learn more from scratch and cannot correct retargeting errors at run time.
\end{itemize}

\textbf{Evaluation Metrics.}
We quantitatively evaluate the performance of all methods with four key metrics: grasping success rate $\text{SR}_{\text{Grasp}}$, following success rate $\text{SR}_{\text{Follow}}$, position error $E_p$ (m) and rotation error $E_r$ (radian). $\text{SR}_{\text{Grasp}}$ reflects successful grasping and lifting that occurs during the whole trajectory. $\text{SR}_{\text{Follow}}$ reflects successful trajectory following where objects must be constantly lifted without falling from the robot hand along the entire trajectory.
Two success rates are counted among the entire test set. $E_p$ and $E_r$ are the average translation and rotation errors between current trajectories and targeted object trajectories. Those two metrics are only counted in successful trajectories after the objects are grasped.
It is worth noticing that In our setup, the episode length mainly follows the human demonstration, so factors like initial human–object distance differ across demos and do not reflect algorithmic speed. To make evaluation fair, we use a fixed time window $T$ a trajectory is counted successful only if it finishes within $T$ and without re-grasps. Under this protocol, all methods have comparable inference time $10\,\mathrm{s}$.

\begin{figure}[t]
  \centering
   \includegraphics[width=0.9\columnwidth]{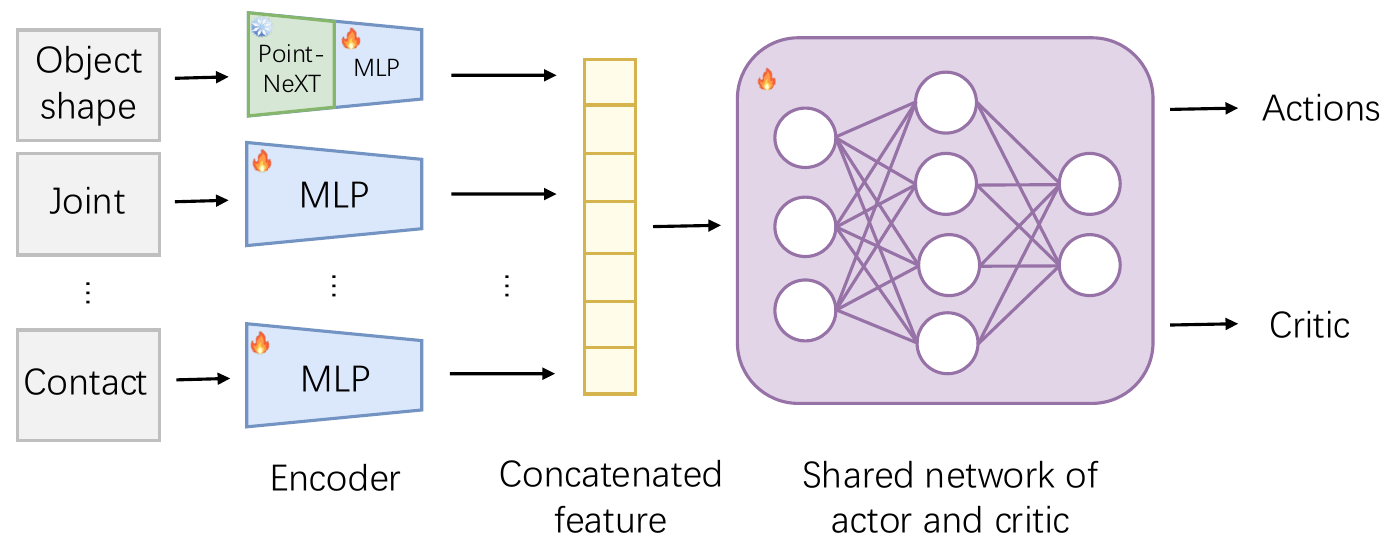}
   \caption{Network structure of DexH2R.}
   \label{fig:network}
\end{figure}

\textbf{Implementation details}
We utilize Isaac Gym~\cite{makoviychuk2021isaac} for simulation. The robot's control frequency is set to 10 Hz with a PD controller for low-level position control.
PPO~\cite{schulman2017proximal} is employed to train the residual policy and only one 3090 GPU is needed for policy training. Network structure of our residual policy is shown in Figure~\ref{fig:network}, where all features after encoders are concatenated and sent into actor-critic policy. For encoding features, all inputs are encoded with MLP except for the object shape, which is encoded with a pretrained PointNeXt~\cite{qian2022pointnext} network to better process point cloud information. Unless otherwise specified, we apply the Leap Hand~\cite{shaw2023leap} with a FANUC robot arm in the experiments. Table~\ref{table:reward} and~\ref{table:tp} shows the detailed setting of reward parameters and training parameters.

The two-stage reward hyperparameters are chosen as follows. (1) In the first stage we jointly constrain fingertip and fingerdip positions to preserve finger directions and improve grasping, but only fingertip position is used for termination/reset since fingerdip alignment is not a hard constraint. (2) In the second stage, lifting and trajectory-following terms get the largest weights because they are the final goals, and the lifting term is linearized to better handle multi-finger coordination. 3) The decay coefficient is set so the exponential term is between $e^{-2}$ and $e^{-1}$ in the first episode before weighting. 4) The reset threshold is set so about half of the environments terminate in the first episode, which stabilizes training.
\begin{table}[t]
\centering
\caption{Reward parameter. Several things to notice: 1) Only exponent designed reward has decay coefficient, while linear reward are labeled as N/A in table. 2) N/A in threshold means there is no threshold. 3) lift up reward has two thresholds. Within this threshold, reward grows linearly based on the lift-up distance.}
\small
\setlength{\tabcolsep}{10pt}
\begin{threeparttable}
\resizebox{\columnwidth}{!}{
\begin{tabular}{c|cccc}
\toprule
Reward category & weight & decay coefficient & threshold\\
\midrule 
\cmidrule(lr){1-1} \cmidrule(lr){2-4}
fingertips & 1.0 & -0.4 & 0.1 \\
fingerdips & 0.5 & -0.4 & N/A \\
velocity smoothness & 0.2 & -5 & 0.5 \\
lift up & 10.0 & N/A & [0.01,0.08] \\
hand reaching objects & 1.0 & -0.5 & 0.07 \\
object following trajectory & 10.0 & -0.5 & 1.0 \\

\bottomrule     
\end{tabular}
}
\end{threeparttable}
\label{table:reward}
\end{table}
\begin{table*}[t]
\centering
\caption{Encoder design and training parameter. Encoder parameter variations is caused by different number of fingers among embodiments.}
\small
\setlength{\tabcolsep}{10pt}
\begin{threeparttable}
\resizebox{\textwidth}{!}{
\begin{tabular}{|c|c|c|c|c|c|c|c|c|c|c|c|c}
\hline 
 & \multicolumn{8}{c|}{Encoder Network} & \multicolumn{3}{c|}{Training setting} \\
 \cline{1-9} \cline{10-12}
 & \makecell{object\\shape} & \makecell{object\\scale} & \makecell{human hand\\trajectory} & \makecell{history\\joint} & \makecell{object\\poses} & \makecell{object follow\\distances} & \makecell{object-hand\\distances} & contact & \makecell{learning\\rate} & \makecell{batch\\size} & \makecell{horizon\\length}\\
  \cline{1-12}
\makecell{LeapHand\\Allegro} & \multirow{2}{*}{[1024,256,64]} & \multirow{2}{*}{[1,64]} & [288,512,256] & [22, 128] & \multirow{2}{*}{[30, 128]} & \multirow{2}{*}{[198, 128]} & [48, 128] & [4, 128] & \multirow{2}{*}{1e-5} & \multirow{2}{*}{32768} & \multirow{2}{*}{32} \\
\cline{1-1}\cline{4-5} \cline{8-9}
Shadow & & & [360,512,256] & [30,128] & & & [72, 128] & [5, 128]& & &\\
\hline     
\end{tabular}
}
\end{threeparttable}
\label{table:tp}
\end{table*}




\subsection{Results and Comparisons} 
\noindent We conduct extensive experiments to evaluate \ours in both simulation and real-robot environments comprehensively. Specifically, we unfold the results of the targeted experiments to address the following key questions: 
\begin{itemize}
    \item \textbf{Q1:} Can \ours improve the retargeting performance? 
    \item \textbf{Q2:} Is \ours generic among different embodiments?
    \item \textbf{Q3:} Can we deploy \ours to a real-world robot?
    \item \textbf{Q4:} Which components are critical to \ours?
    \item \textbf{Q5:} What is the current limitation of \ours?
\end{itemize}
\begin{table*}[t]
\caption{Quantitative results for seen objects, unseen trajectories and unseen objects with our method.
}
\centering
\setlength{\tabcolsep}{8pt}
\begin{threeparttable}
\resizebox{\textwidth}{!}{
\begin{tabular}{c|c|cccc|cccc|cccc}
\toprule
\multicolumn{2}{c}{}  & \multicolumn{4}{|c}{Leap Hand} & \multicolumn{4}{|c}{Allegro Hand} & \multicolumn{4}{|c}{Shadow Hand}\\
\midrule
Test set & Method & $\text{SR}_{\text{Grasp}}$ $\uparrow$ & $\text{SR}_{\text{Follow}}$ $\uparrow$ & $E_p$ $\downarrow$ & $E_r$ $\downarrow$ & $\text{SR}_{\text{Grasp}}$ $\uparrow$ & $\text{SR}_{\text{Follow}}$ $\uparrow$ & $E_p$ $\downarrow$ & $E_r$ $\downarrow$ & $\text{SR}_{\text{Grasp}}$ $\uparrow$ & $\text{SR}_{\text{Follow}}$ $\uparrow$ & $E_p$ $\downarrow$ & $E_r$ $\downarrow$ \\ 
\midrule

\multirow{6}{*}{\makecell{Seen\\objects}} 
& Position & 32.0\% & 13.7\% & 0.083 & \textbf{0.364} & 25.0\% & 12.3\% & 0.077 & \textbf{0.414} & 17.4\% & 5.9\% & 0.103 & 0.675  \\
& Vector & 32.7\% & 9.1\% & 0.108 & 0.649 & 25.3\% & 6.0\% & 0.096 & 0.632 & 22.5\% & 5.3\% & 0.131 & 0.884\\
& Dexpilot & 32.8\% & 10.7\% & 0.111 & 0.523 & 23.3\% & 6.2\% & 0.092 & 0.578 & 22.5\% & 5.3\% & 0.131 & 0.884 \\
& GeoRT & 32.4\% & 12.3\% & 0.055 & 0.245& 23.7\% & 7.3\% & 0.093 & 0.575 & 31.1\% & 8.1\% & 0.118 & 0.557 \\
& ViViDex-SP &44.6\% &20.0\% & \textbf{0.038} & 0.278 & 28.9\% & 7.9\% & 0.084 & 0.501 & 30.6\% & 8.7\% & 0.110 & \textbf{0.508}  \\
& \textbf{Ours} & \textbf{69.8\%} &\textbf{52.1\%} & 0.048 & 0.452 & \textbf{69.8\%} &\textbf{52.5\%} & \textbf{0.054} & 0.483 & \textbf{65.4\%} &\textbf{43.4\%} & \textbf{0.084} & 0.616\\
\midrule
\multirow{6}{*}{\makecell{Unseen\\objects}} 
& Position & 29.0\% & 14.3\% & 0.091 & 0.389 & 23.9\% & 13.0\% & 0.074 & \textbf{0.375} & 13.4\% & 7.6\% & 0.124 & 0.568 \\
& Vector & 31.8\% & 10.2\% & 0.103 & 0.533 & 22.3\% & 8.7\% & 0.083 & 0.479 & 16.2\% & 5.9\% & 0.115 & 0.829\\
& Dexpilot & 32.2\% & 11.0\% & 0.105 & 0.545 & 22.5\% & 7.2\% & 0.083 & 0.539 & 15.9\% & 7.6\% & 0.117 & 0.692\\
& GeoRT & 29.9\% & 9.7\% & 0.063 & \textbf{0.221} & 19.9\% & 5.1\% & 0.095 & 0.449 & 24.6\% & 4.9\% & 0.105 & \textbf{0.488} \\
& ViViDex-SP & 33.1\% & 14.4\% & \textbf{0.043} & 0.309 & 21.5\% & 5.3\% & 0.106 & 0.652 & 17.0\% & 4.4\% & 0.103 & 0.500 \\
& \textbf{Ours}  & \textbf{70.9\%} &	\textbf{52.7\%} & 0.055 & 
0.476 & \textbf{69.1\%} &	\textbf{53.5\%} & \textbf{0.055} & 
0.568 & \textbf{62.4\%} &	\textbf{46.9\%} & \textbf{0.103} & 
0.808 \\
\bottomrule
\end{tabular}
}
\end{threeparttable}
\label{table:simulation}
\end{table*}
\begin{figure}[t]
  \centering
   \includegraphics[width=\columnwidth]{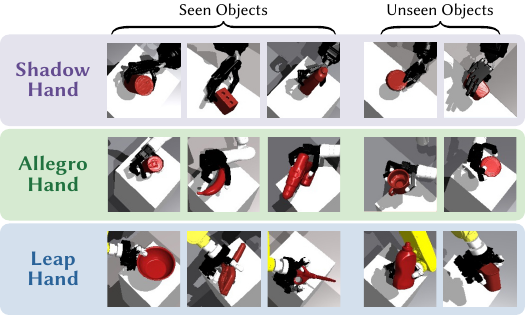}
   \caption{Simulation results showing that \ours can generalize to both seen and unseen objects even for those with long-tailed and rare shapes in three different embodiments.}
   \label{fig:simulation}
\end{figure}
\begin{figure}[t]
  \centering
   \includegraphics[width=0.8\columnwidth]{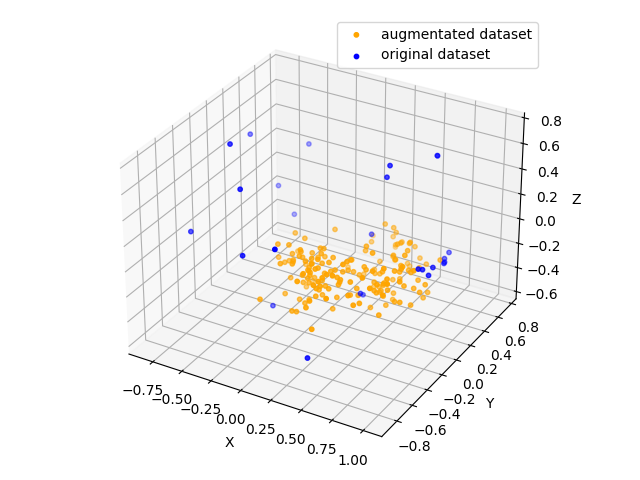}
   \caption{Data distribution of original and augmented dataset.}
   \label{fig:all}
\end{figure}
\subsection*{\textbf{Q1:} Can \ours improve the retargeting performance?}
\noindent In Table~\ref{table:simulation}, we report the quantitative results in the simulation environment. It is obvious that the retargeting baselines achieve poor performance when there is no real-time human correction. Without the capacity to adapt to task dynamics or environmental feedback, these baselines often fail to maintain stable manipulation, especially during long-horizon execution. Additionally, data-driven method also underperforms, which further confirms that our strategy of leveraging retargeted demonstrations as primitive actions, together with our reward design, is effective. \ours significantly improves the success rates of both grasping and trajectory following, showing a clear advantage in terms of robustness and adaptability. Especially, notable performance gains are witnessed in the $SR_{\text{follow}}$ metric, which reflects the great task completion performance of \ours. Specifically, \ours can follow desired trajectories throughout the entire manipulation process without falling, which is crucial for practical applications with high robustness requirements. 
These improvements suggest that our method is capable of producing temporally consistent and stable behaviors by remaining contact between robots and objects, which demonstrates that our method effectively bridges the gap between human demonstrations and dexterous robot execution, ensuring successful task completion across diverse scenarios. 

We also provide some visualizations in Figure~\ref{fig:simulation}, where \ours can grasp both seen and unseen objects properly including those with complex geometrical shapes. Such results highlight the necessity of our proposed task-oriented residual policy, which takes the task-specific goal and environment information into consideration. This design enables the policy to focus on critical aspects of the manipulation task, such as contact stability and path precision, ultimately contributing to its generalizability and robustness.

It is worth mentioning that some baselines are comparable with DexH2R in tracking errors because they can only follow some trivial trajectories, and only successful trajectory-following cases are considered in the calculation of tracking errors. 
Specifically, Position and orientation tracking errors $(E_p, E_r)$ are computed only on trajectories that finish successfully (i.e., the object does not slip or fall during the entire episode). Because our method achieves a much higher finishing success rate $\mathrm{SR}_{\text{Follow}}$, the set over which its errors are averaged is larger and contains more challenging cases (e.g., longer motions or larger pose changes), whereas several baselines contribute fewer and typically easier successful episodes. Nevertheless, improving tracking accuracy remains an important goal, and we will continue to reduce errors while maintaining high success.

\subsection*{\textbf{Q2:} Is \ours generic among different embodiments?}
\noindent We claim \ours as a generic algorithm compatible with different embodiments with minimal hyperparameter adjustments during policy training. In this part, we implement our policy in three distinct embodiments: the Leap Hand with a FANUC robot arm, the Allegro Hand with a Kinova robot arm, and the Shadow Hand with a UR10e robot arm. These embodiments vary significantly in terms of joint configurations and physical dynamics, making it a challenging testbed for generalization. 

Despite these differences, quantitative results in Table~\ref{table:simulation} indicate that our policy performs greatly in all of them and significantly outperforms baselines in both grasp success and trajectory-following metrics. Such consistency across embodiments suggests that \ours is robust to variations in kinematic structures and control latency, which are common challenges in real-world multi-robot deployment. This demonstrates the adaptability of our policy across robot systems as a generic solution that does not require architecture-specific re-design. Furthermore, our policy enables direct successful policy training in new hardware with only minimal hyperparameter adjustment, reducing the tuning effort traditionally required when adapting learned models to different robots. This highlights a significant advantage: by leveraging \ours, a single dataset of human demonstrations can be successfully transferred to various robot embodiments for task execution without the need for hardware-specific optimization, substantially enhancing the efficiency of robot data collection.


\subsection*{\textbf{Q3:} Can we deploy \ours to a real-world robot?}
\noindent We deploy our policy to a real robot system composed of a FANUC LR Mate robot with a Leap Hand~\cite{shaw2023leap}. We evaluated our policy on four different objects, including two seen and two unseen from the YCB dataset~\cite{calli2015ycb}. Objects are scaled up to match their sizes in human demonstrations. We conduct 10 experiments for each object through open-loop execution and record the success rates of grasping $\text{SR}_{\text{Grasp}}$ and trajectory following $\text{SR}_{\text{Follow}}$. The evaluation covers both seen and unseen objects with diverse poses and orientations, providing insight into the generalizability of \ours across different scenarios.

As shown in Table~\ref{table:real_world}, our proposed method successfully bridges the sim-to-real gap by achieving relatively high performance in the success rates of both grasping and trajectory following. In most cases, the hand pose remains fairly consistent in the entire manipulation process, demonstrating its practical feasibility. We also notice a drop in the success rate of trajectory following for unseen objects due to the lack of closed-loop feedback and object-wise physical properties especially if the unseen objects have irregular shapes or the desired trajectories contain complex rotations.

\begin{figure*}[t]
  \centering
   \includegraphics[width=\textwidth]{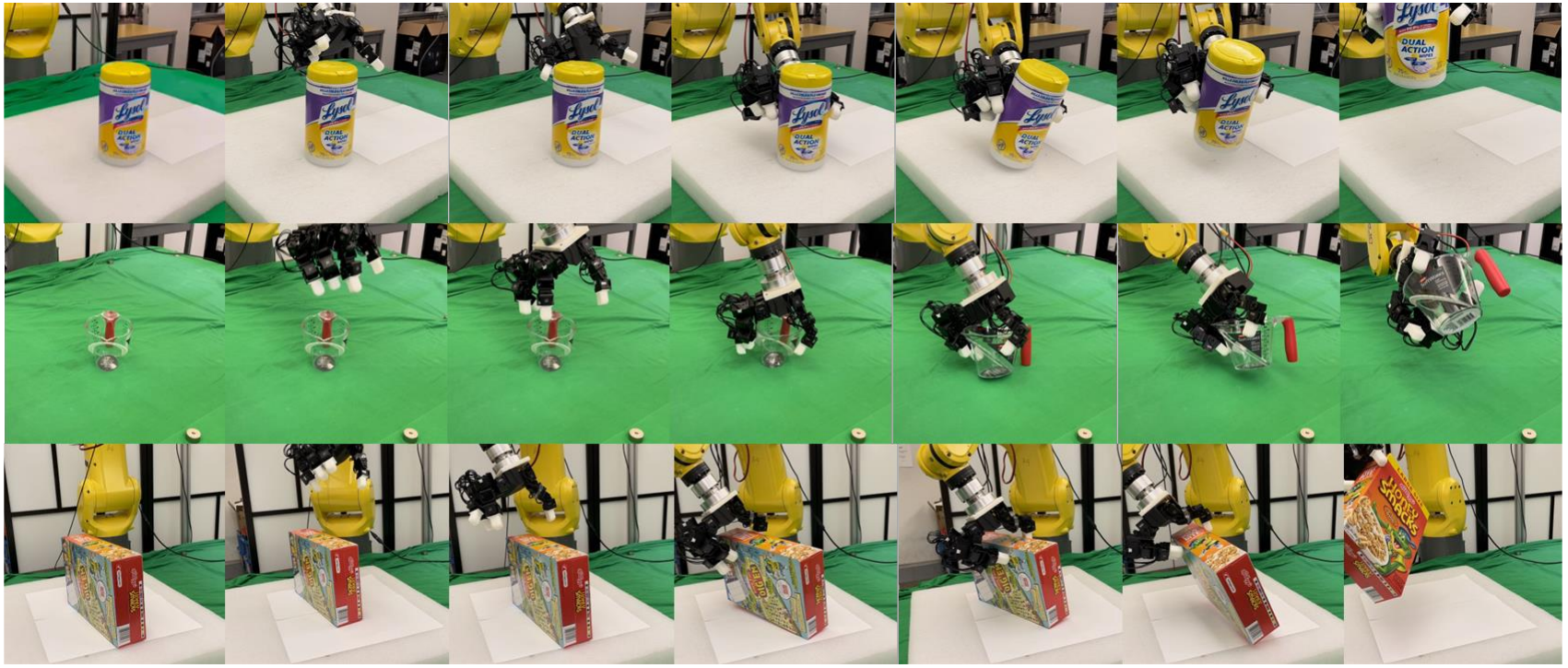}
   \caption{Real-world experiments. \ours can generalize to various objects with each row as a manipulation action sequence. }
   \label{fig:real_world}
\end{figure*}

We visualize some qualitative results in Figure~\ref{fig:real_world}, which shows that \ours can successfully grasp the objects and follow the desired trajectories with diverse objects in the real world. These visualizations further validate the adaptability of our approach and its potential for practical deployment in unstructured environments.

\begin{table}[t]
\caption{Performance of our method on different objects.}
\small
\setlength{\tabcolsep}{10pt}
\begin{threeparttable}
\resizebox{\columnwidth}{!}{
\begin{tabular}{c|c|cccc}
\toprule
\multicolumn{2}{c}{Objects} & $\text{SR}_{\text{Grasp}}$ $\uparrow$ & $\text{SR}_{\text{Follow}}$ $\uparrow$ & $E_p$ $\downarrow$ & $E_r$ $\downarrow$ \\ 
\midrule
\multirow{5}{*}{\makecell{Seen\\objects}} & master chef can & 84.7\%  &	63.3\% & 0.035 & 0.259 \\
& bleach cleanser & 87.5\% & 75.9\% & 0.031 & 0.301 \\
& scissors & 31.2\% & 29.6\% & 0.066 & 0.598  \\
& power drill  & 65.8\% & 34.8\% & 0.05 & 0.408 \\
& Overall  &  \textbf{69.8\%} &\textbf{52.1\%} & \textbf{0.048} & \textbf{0.452} \\
\midrule
\multirow{3}{*}{\makecell{Unseen\\objects}}& potted meat can & 71.9\% & 57.3\% & 0.045 & 0.496 \\
& mug & 62.0\% & 47.0\% & 0.062 & 0.489 \\
& Overall  &  \textbf{70.9\%} &	\textbf{52.7\%} & \textbf{0.055} & \textbf{0.476} \\

\bottomrule     
\end{tabular}
}
\end{threeparttable}
\label{table:ablation_obj}
\end{table}
\begin{table}[t]
\caption{Ablation studies of two different stages. Experiments are conducted on unseen objects to make the differences more clear.}
\centering
\setlength{\tabcolsep}{10pt}

\begin{minipage}[t]{0.48\textwidth}
\caption*{a) Ablation studies on staging pipeline.}
\centering
\resizebox{\linewidth}{!}{
\begin{tabular}{ccccc}
\toprule
Method & $\text{SR}_{\text{Grasp}}$ $\uparrow$ & $\text{SR}_{\text{Follow}}$ $\uparrow$ & $E_p$ $\downarrow$ & $E_r$ $\downarrow$ \\ 
\cmidrule(lr){1-1} \cmidrule(lr){2-5}
Ours & \textbf{70.9\%} & \textbf{52.7\%} & \textbf{0.055} & \textbf{0.476} \\
w/o data aug & 54.7\% & 34.1\% & 0.079 & 0.557 \\
w/o prim actions & 66.3\% & 47.4\% & 0.055 & 0.553 \\
\bottomrule
\end{tabular}
}
\label{table:pipeline}
\end{minipage}
\hfill
\begin{minipage}[t]{0.48\textwidth}
\vspace{5pt}
\caption*{b) Ablation studies on residual policy.}
\centering
\resizebox{\linewidth}{!}{
\begin{tabular}{ccccc}
\toprule
Method & $\text{SR}_{\text{Grasp}}$ $\uparrow$ & $\text{SR}_{\text{Follow}}$ $\uparrow$ & $E_p$ $\downarrow$ & $E_r$ $\downarrow$ \\
\cmidrule(lr){1-1} \cmidrule(lr){2-5}
Ours & \textbf{70.9\%} & \textbf{52.7\%} & \textbf{0.055} & \textbf{0.476} \\
w/o contacts & 67.6\% & 47.5\% & 0.056 & 0.511 \\
w/o dense reward & 17.4\% & 2.7\% & 0.094 & 0.487 \\
\bottomrule
\end{tabular}
}
\label{table:residual}
\end{minipage}

\label{table:ablation}
\end{table}
\begin{table}[t]
\caption{DexH2R with different switching steps.}
\small
\setlength{\tabcolsep}{10pt}
\begin{threeparttable}
\centering
\resizebox{\columnwidth}{!}{
\begin{tabular}{c|cccc}
\toprule
Switching step & $\text{SR}_{\text{Grasp}}$ $\uparrow$ & $\text{SR}_{\text{Follow}}$ $\uparrow$ & $E_p$ $\downarrow$ & $E_r$ $\downarrow$ \\ 
\midrule
 5 & 64.9\%  & 46.8\% & 0.041 & 0.431 \\
 10 & 68.8\% & 52.0\% & 0.040 & 0.430 \\
 15 & 69.8\% & 52.1\% & 0.048 & 0.452  \\
 20 & 65.7\% & 48.7\% & 0.048 & 0.420 \\
 25 & 64.3\% & 44.9\% & 0.049 & 0.446 \\

\bottomrule     
\end{tabular}
}
\end{threeparttable}
\label{table:ablation}
\end{table}
\begin{figure}[t]
  \centering
   \includegraphics[width=\columnwidth]{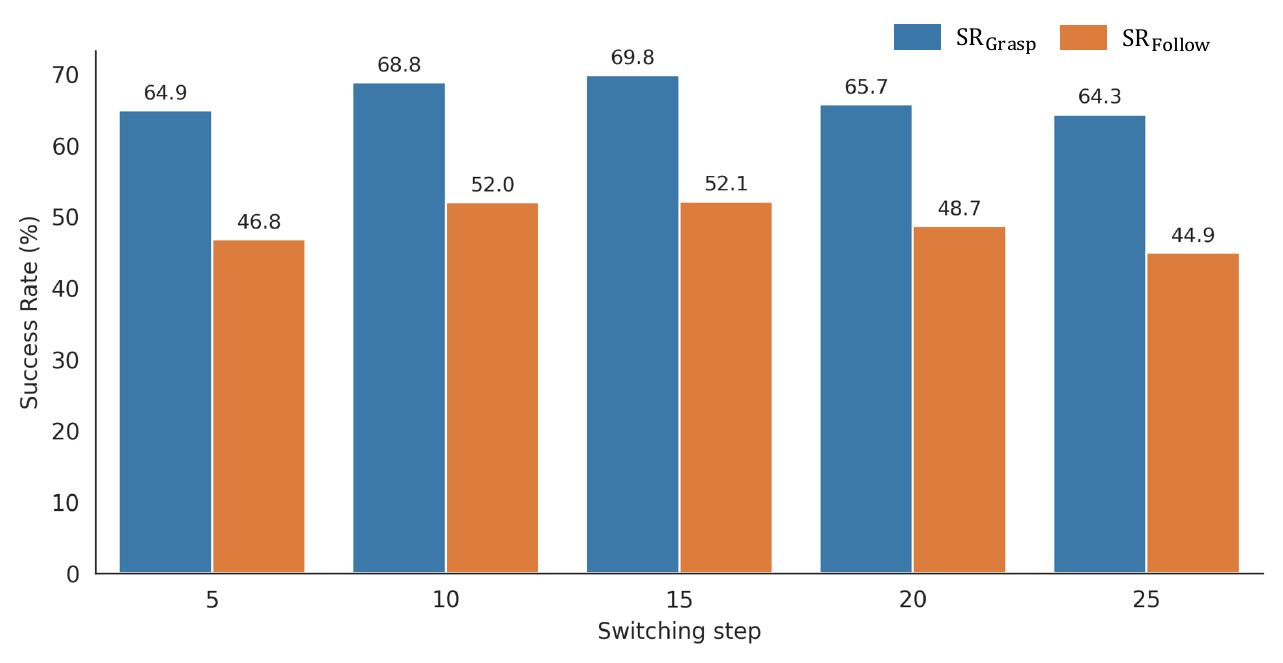}
   \caption{DexH2R with different switching steps.}
   \label{fig:switching}
\end{figure}

\begin{table}[t]
\caption{Success rates of real-world experiments on both seen and unseen objects.}

\centering
\small
\setlength{\tabcolsep}{10pt}
\begin{threeparttable}
\begin{tabular}{c|c|cc}
\toprule
\multicolumn{2}{c}{Objects} & $\text{SR}_{\text{Grasp}}$ $\uparrow$ & $\text{SR}_{\text{Follow}}$ $\uparrow$ \\ 
\midrule
\multirow{2}{*}{\makecell{Seen\\objects}} & sugar box & 10/10 &	9/10 \\
& tomato soup can & 9/10 & 7/10 \\
\midrule
\multirow{2}{*}{\makecell{Unseen\\objects}} 
& mug & 8/10 & 5/10 \\
& mustard bottle & 8/10 & 4/10  \\

\bottomrule     
\end{tabular}
\end{threeparttable}
\label{table:real_world}
\end{table}


\begin{table}[t]
\caption{Training time and converge iterations for our policies and policy without primitive actions.}
\centering
\setlength{\tabcolsep}{10pt}

\begin{minipage}[t]{0.48\textwidth}
\caption*{a) Time performance of our method across different embodiments.}
\centering
\resizebox{\linewidth}{!}{
\begin{tabular}{c|cc}
\toprule
Policy & Training Time & Iterations for converge \\ 
\midrule
 LeapHand & 16.53h & 2593 \\
 Allegro & 21.98h & 3504 \\
 Shadow & 21.98h & 3504 \\
\bottomrule
\end{tabular}
}
\label{table:timea}
\end{minipage}
\hfill
\begin{minipage}[t]{0.48\textwidth}
\vspace{5pt}
\caption*{b) Time performance comparison between our method with and without primitive actions, where 'w prim' means with primitive actions and 'w/o prim' means without.}
\centering
\resizebox{\linewidth}{!}{
\begin{tabular}{ccccc}
\toprule
Policy & Training Time & Iterations for converge \\ 
\midrule
 w prim & 16.53h & 2593 \\
 w/o prim & 21.98h & 3504 \\
\bottomrule
\end{tabular}
}
\label{table:timeb}
\end{minipage}

\label{table:time}
\end{table}
\subsection*{\textbf{Q4:} Which components are critical to \ours?}
\noindent In this part, we investigate the role of each module in \ours. Specifically, we dig into the necessity of each stage in our pipeline in Table~\ref{table:pipeline} a). A significant performance drop is witnessed if we do not apply data augmentation in human data extraction since data augmentation can greatly boost the diversity of demonstrations in human data. This is because the augmented dataset covers a larger portion of the workspace. Figure~\ref{fig:all} compares the original and augmented pose distributions of object 'master chef can', increasing number of data in workspace from \textbf{14} to \textbf{182}. Besides, the elimination of the primitive action slightly hurts the model performance. Although reinforcement learning can derive a reasonable policy even without primitive actions, it would take a long time to converge (see Table~\ref{table:time}) and can hardly ensure the smoothness of the trajectories. This underscores the benefit of primitive actions in speeding up the learning process, as they provide a good initialization for the policy. 

In Table~\ref{table:residual} b), we analyze the design of the residual in the second stage. The binary contact information in the observation can provide useful cues about the interaction between human hands and objects, which helps to lift the performance. We also emphasize the importance of the dense rewards. Without the guidance provided by human hand rewards, the exploration of the residual policy becomes almost unmanageable due to the high-dimensional action space, which aligns with the challenges described in~\cite{dasari2023learning}. This highlights the necessity of our two-stage reward to provide directional guidance during training, especially when explicit supervision is limited.

We also conducted experiments to assess how object properties affect performance. As demonstrated in Table~\ref{table:ablation_obj}, the dexterous hand achieves higher success rates with objects that have regular shapes and suitable sizes. These objects present more feasible grasping solutions compared to objects with irregular shapes or extreme sizes because the consistent geometry and contact surface of these regular objects allow for more stable and repeatable contact interactions during manipulation. This finding highlights the importance of object properties in determining the effectiveness of manipulation tasks, suggesting a promising direction for future research.
Additionally, the pre-defined switching step is initially selected based on prior experience and we also conducted an ablation over different switching steps shown in Table~\ref{table:ablation} and Figure~\ref{fig:switching}. The quantitative results indicate that, within a reasonable range, the switching step has limited impact on final performance.

\subsection*{\textbf{Q5:} What is the current limitation of \ours?}
\noindent Although DexH2R improves success rates over retargeting baselines, we observe several method-specific failure modes that manifest consistently across simulation and real hardware.

\textbf{Long-horizon drift without closed-loop correction.} Despite error mitigation brought our residual design, small pose errors and contact noise can still accumulate over extended trajectories, leading to late-stage object drops or poses offsets during precise rotations. 

\textbf{Physics- and shape-dependent fragility.} Irregular geometries, scales, and physical parameters of object increase sensitivity to small pose deviations. The effect is more pronounced on unseen objects and in sequences requiring sustained rotational alignment because of unseen shape and physical parameters. This is also amplified in the real robot experiments, where more out-of-distribution environment settings are involved such as friction, mass and contact.  

\textbf{Perception mismatch} Because DexH2R conditions on human-hand keypoints and object encodings, their perception inaccuracies can shift the intended contact region at approach, elevating failure risk near first touch. Even though our reward design can mitigate such bad effect by focusing on object contact at second stage, these inaccuracies still reduce policy’s sensitivity to this input channel, thereby degrading overall accuracy.

Overall, despite strong performance, DexH2R can still struggle on precise, long-horizon rotations without feedback, during contact-rich transitions on irregular objects, and under perception noise of human hand and object. These limitations arise from open-loop execution, lack of training data, and sensitivity to perception mismatch, suggesting promising directions for future work.

\section{Conclusions}
\noindent In this work, we introduce DexH2R, a framework that transfers human hand trajectories to dexterous robot motions while achieving task completion without on-time human feedback.
By decoupling our policy execution from human intervention, the system is more suitable for autonomous deployment when ensuring the successful task completion. 
Also, \ours is generic and compatible with different embodiments without hardware-specific adjustment for human demonstrations. Those two benefits enable \ours to possess large potential in becoming a scalable data collection method, lowering the demand on both hardware and human labor. With comprehensive experiments on three different embodiments, our method exhibits a notable performance gain compared with baselines on grasping tasks in both simulation and real-world environments.

In the future, we plan to extend DexH2R to more complex tasks, such as in-hand manipulation, to further test its generality under richer contact interactions. Another direction is integrating tactile sensing to provide fine-grained contact information, enhancing the policy’s perception of subtle contact dynamics and enabling more precise manipulation in contact-rich scenarios. We also envision exploring improved training protocols and evaluation suites that stress long-horizon stability and robustness across diverse objects and embodiments.

\bibliographystyle{IEEEtran}
\bibliography{ref}

 


\vspace{-30pt}
\begin{IEEEbiography}[{\includegraphics[width=0.8in,height=1.0in,clip,keepaspectratio]{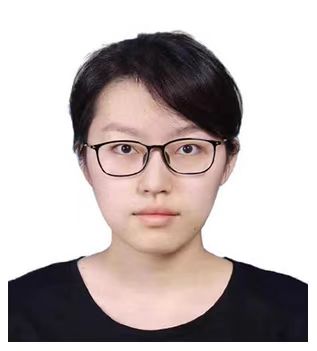}}]{Shuqi Zhao}
is currently a Ph.D. student in Mechanical Engineering from UC Berkeley, advised by professor Masayoshi Tomizuka. She received the B.S. degree in Control Science and Engineering with an honor minor  degree in Chu Kochen Honor College from Zhejiang University in 2023 and her research interests include dexterous manipulation, reinforcement learning, and sim-to-real transfer.
\end{IEEEbiography}
\vspace{-50pt}
\begin{IEEEbiography}[{\includegraphics[width=0.8in,height=1.0in,clip,keepaspectratio]{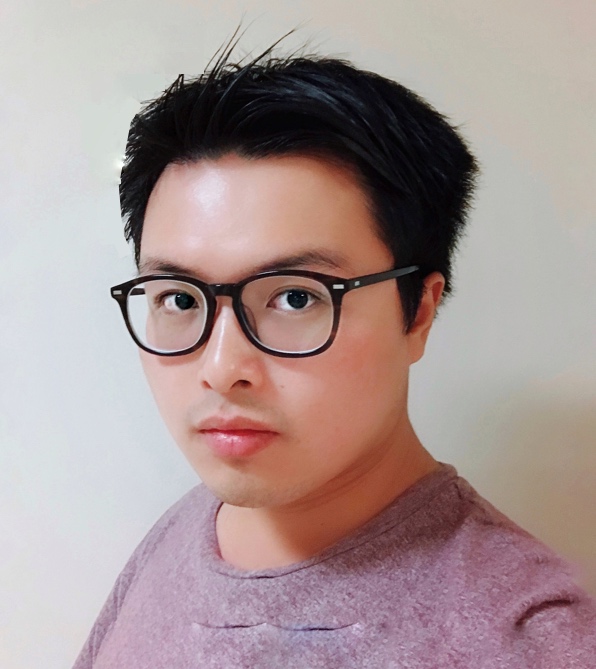}}]{Xinghao Zhu}
holds a B.E. in Automation from Xi'an Jiaotong University (2018) and a Ph.D. in Mechanical Engineering from the University of California, Berkeley (2024). His research interests span control theory, robotics, machine learning, and optimization. He primarily investigates problems in contact-rich and dexterous manipulation, as well as loco-manipulation.
\end{IEEEbiography}
\vspace{-50pt}
\begin{IEEEbiography}[{\includegraphics[width=0.8in,height=1.0in,clip,keepaspectratio]{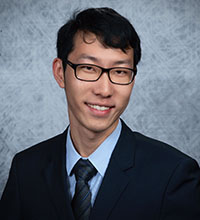}}]{Yuxin Chen}
received his B.Sc. degree in Mechanical Engineering from Shanghai Jiao Tong University, Shanghai, China, in 2020, and both the B.S.E. degree in Aerospace Engineering and the M.S. degree in Robotics from the University of Michigan, Ann Arbor, MI, USA, in 2020 and 2022, respectively. He is currently pursuing a Ph.D. in Mechanical Engineering at the University of California, Berkeley.
His research interests lie in control, optimization, and reinforcement learning, with a focus on mobile robotics, dexterous manipulation, and human–robot interaction.
\end{IEEEbiography}
\vspace{-50pt}
\begin{IEEEbiography}[{\includegraphics[width=0.8in,height=1.0in,clip,keepaspectratio]{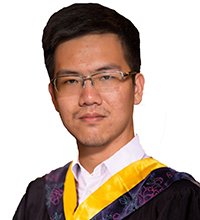}}]{Chenran Li}
is currently a researcher scientist in Nvidia. He received his Ph.D. degree from University of California, Berkeley in 2025 and his B.S. degree from Harbin Institute of Technology in 2020. His research interests include humanoid and reinforcement learning.
\end{IEEEbiography}
\vspace{-40pt}
\begin{IEEEbiography}[{\includegraphics[width=0.8in,height=1.0in,clip,keepaspectratio]{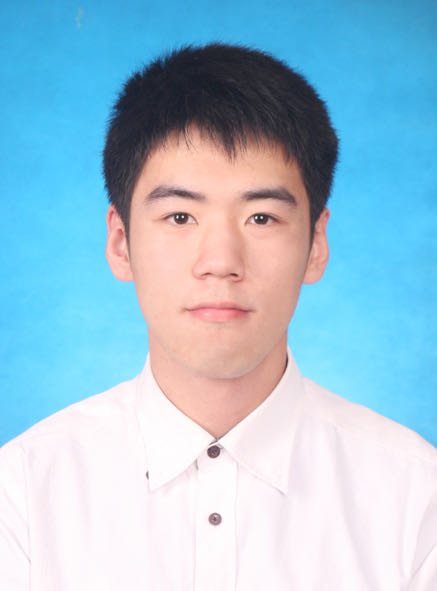}}]{Yichen Xie}
received his B.E. degree in Computer Science from Shanghai Jiao Tong University in 2021. He is currently pursuing the Ph.D. degree in Mechanical Engineering at University of California, Berkeley, advised by Prof. Masayoshi Tomizuka. His research interest covers computer vision and robotics. He has published papers at conferences such as CVPR, ICCV, ICLR, and NeurIPS, and has served as a reviewer for CVPR, ECCV, NeurIPS, ICLR, and ICML.
\end{IEEEbiography}
\vspace{-30pt}
\begin{IEEEbiography}[{\includegraphics[width=0.8in,height=1.0in,clip,keepaspectratio]{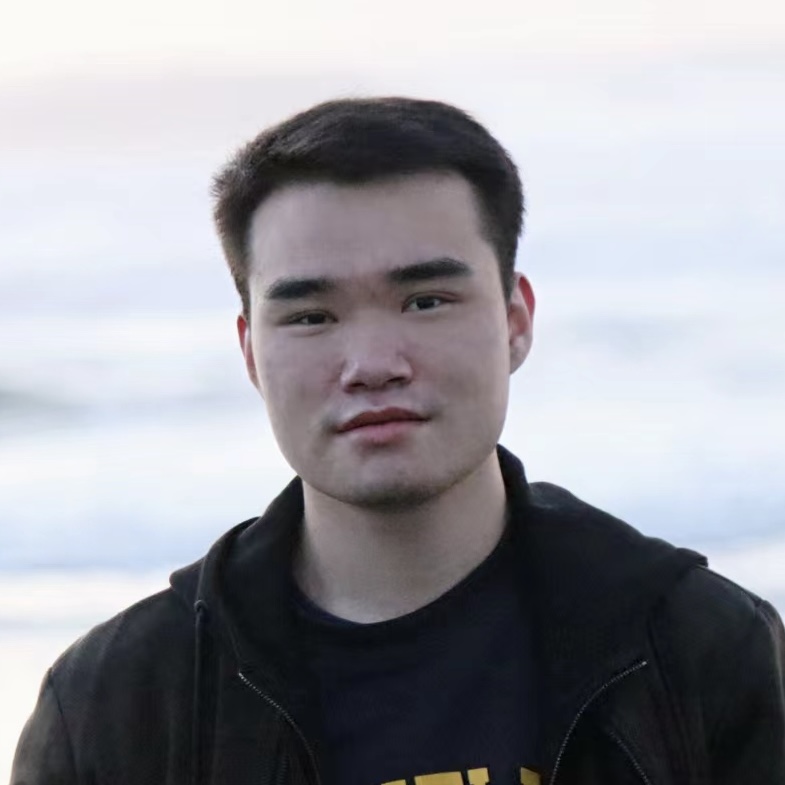}}]{Xiang Zhang}
received the B.Eng. degree in pricision mechnical and instrument from the University of Science and Tech- nology of China, Hefei, China, in 2019 and the Ph.D. degree in mechanical and control engineering from the University of California, Berkeley, CA, USA, in 2024.
He is currently with FANUC Advanced Research Laboratory, FANUC America Corporation, USA. His research interests include robot modeling, control, and manipulation.
\end{IEEEbiography}
\vspace{-30pt}
\begin{IEEEbiography}[{\includegraphics[width=0.8in,height=1.0in,clip,keepaspectratio]{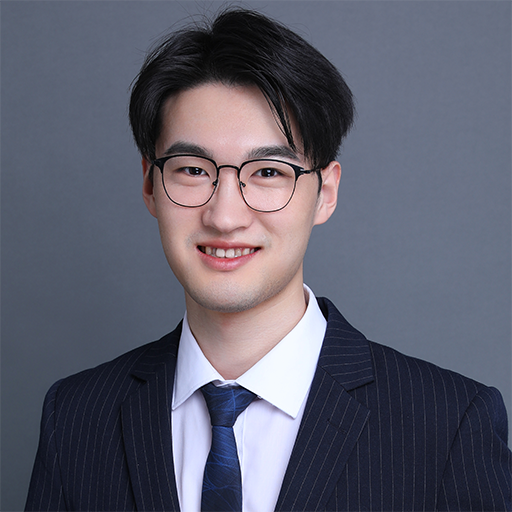}}]{Mingyu Ding}
is an assistant professor of computer science at the University of North Carolina at Chapel Hill. Prior to joining the department, he was a postdoctoral fellow at BAIR@UC Berkeley with Masayoshi Tomizuka and a visiting scholar at CSAIL@MIT with Joshua Tenenbaum. He obtained a doctorate from the University of Hong Kong in 2022 with Ping Luo. Mingyu’s research interests lie at the intersection of robotics, embodied AI, and vision. He received Baidu Fellowship, Microsoft Fellowship Nomination, ME Rising Star, WAIC Rising Star, and the Best Paper Award from ICRA 2024.
\end{IEEEbiography}
\vspace{-25pt}
\begin{IEEEbiography}[{\includegraphics[width=0.8in,height=1.0in,clip,keepaspectratio]{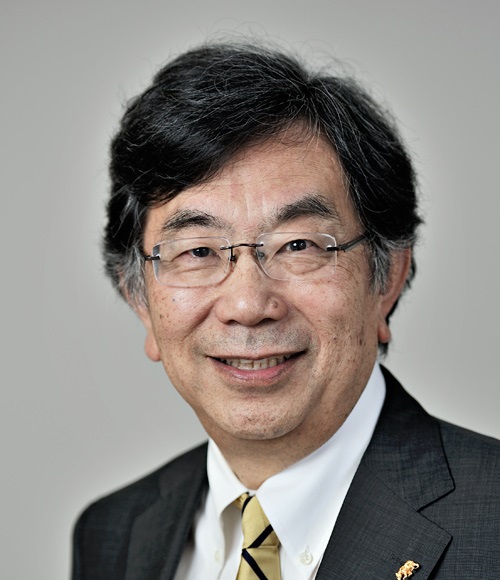}}]{Masayoshi Tomizuka}
was born in Tokyo, Japan, in 1946. He received the B.S. and M.S. degrees in mechanical engineer- ing from Keio University, Tokyo, in 1968 and 1970, respectively, and the Ph.D. degree in me- chanical engineering from the Massachusetts Institute of Technology, Cambridge, MA, USA, in 1974.
In 1974, he joined the Faculty of the De- partment of Mechanical Engineering, University of California at Berkeley, Berkeley, CA, USA,
where he is currently the Cheryl and John Neerhout, Jr., Distinguished Professorship Chair and teaches courses in dynamic systems and con- trols. From 2002 to 2004, he was the Program Director of the Dynamic Systems and Control Program of the Civil and Mechanical Systems Division of NSF. His current research interests include optimal and adaptive control, digital control, signal processing, motion control, and control problems related to robotics, machining, manufacturing, informa- tion storage devices, and vehicles.Dr. Tomizuka was the Technical Editor of the ASME Journal of Dynamic Systems, Measurement and Control (J-DSMC) (1988–1993), Editor-in-Chief of IEEE/ASME TRANSACTIONS ON MECHATRONICS (1997–1999), and an Associate Editor for the Journal of the International Federation of Automatic Control, and Automatica. He was the General Chairman of the 1995 American Control Confer- ence, and was the President of the American Automatic Control Council (1998–1999). He is a Life Fellow of the ASME and a Fellow of the International Federation of Automatic Control (IFAC) and the Society of Manufacturing Engineers. He was the recipient of the Best J-DSMC Best Paper Award (1995 and 2010), DSCD Outstanding Investigator Award (1996), Charles Russ Richards Memorial Award (ASME, 1997), Rufus Oldenburger Medal (ASME, 2002), John R. Ragazzini Award (AACC, 2006), Richard E. Bellman Control Heritage Award (AACC, 2018), Honda Medal (ASME, 2019), and Nathaniel B. Nichols Medal (IFAC, 2020). He is a member of the National Academy of Engineering.
\end{IEEEbiography}

\end{document}